\documentclass[runningheads]{llncs}
\usepackage[numbers]{natbib}
\usepackage{natbib}
\usepackage{float}
\usepackage{mathtools}

\usepackage{multirow}
\usepackage{makecell}
\usepackage{subcaption}

\usepackage{amssymb}
\usepackage[draft]{fixme}
\usepackage{soul}
\usepackage{amsmath}
\usepackage{hyperref}
\usepackage{graphicx}

\DeclarePairedDelimiter\ceil{\lceil}{\rceil}
\DeclarePairedDelimiter\floor{\lfloor}{\rfloor}

\begin{document}
\title{ADNet: Temporal Anomaly Detection in Surveillance Videos}
%
%
\author{Halil İbrahim Öztürk\inst{1, 2} \and
Ahmet Burak Can\inst{1}}
%
%
\institute{Department of Computer Engineering, Hacettepe University, Ankara, Turkey \\
\email{\{halil\_ozturk, abc\}@hacettepe.edu.tr}
\and
Havelsan, Ankara, Turkey \\
\email{hozturk@havelsan.com.tr}\\
}
\maketitle              
\begin{abstract}
Anomaly detection in surveillance videos is an important research problem in computer vision. In this paper, we propose ADNet, an anomaly detection network, which utilizes temporal convolutions to localize anomalies in videos. The model works online by accepting consecutive windows of video clips.  Features extracted from video clips in a window are fed to ADNet, which allows to localize anomalies in videos effectively. We propose the AD Loss function to improve abnormal segment detection performance of ADNet. Additionally, we propose to use F1@k metric for temporal anomaly detection. F1@k is a better evaluation metric than AUC in terms of not penalizing minor shifts in temporal segments and punishing short false positive temporal segment predictions. Furthermore, we extend UCF Crime \cite{sultani2018real} dataset by adding two more anomaly classes and providing temporal anomaly annotations for all classes. Finally, we thoroughly evaluate our model on the extended UCF Crime dataset. ADNet produces promising results with respect to F1@k metric. Code and dataset extensions are publicly at \url{https://github.com/hibrahimozturk/temporal_anomaly_detection} 

\end{abstract}
\section{Introduction}
\label{sec:intro} 

In today's physical world, surveillance systems are placed everywhere to provide security, such as shopping malls, offices, homes, rail stations. Detection of abnormal situations in surveillance systems is crucial to make early intervention possible e.g., helping people harmed by a traffic accident, explosion, shooting. Sometimes preventing growth of undesired events like panic, fight, or fire can be possible by detecting anomalies timely in surveillance video streams. However, analyzing video streams in real-time and detecting abnormal cases require excessive human resources and prone to errors due to lose of human attention with time. Since human observation is not an effective solution, automatic anomaly detection approaches that leverage artificial intelligence mechanisms are needed in surveillance systems.   

Detecting abnormal cases in real world videos is a complex problem, because it is hard to define abnormal cases objectively. Abnormality definition changes according to context or environment, such as seeing a bicycling man in a building can be abnormal while it is not on a road. However, some events can usually be considered as abnormal in most environments like panic in a crowd, fighting persons, vandalism, explosion, fire, etc. Therefore, most anomaly detection studies \cite{kratz2009anomaly,mahadevan2010anomaly, li2013anomaly, yuan2014online,sultani2018real} usually focus on detecting these kind of events. General approach in these studies is to learn normal behavior in the environment and define everything out of the normal as abnormal. This approach might produce some false positives but has potential to recognize anomaly cases that are not included in the training data set. Other possible approach is to teach anomaly classes in a data set to the machine learning model and try to recognize them. In this way the approach becomes an action recognition/localization solution and loses ability to recognize unseen anomaly classes.

\begin{figure}
\begin{center}
\begin{tabular}{c}
\includegraphics[width=\linewidth]{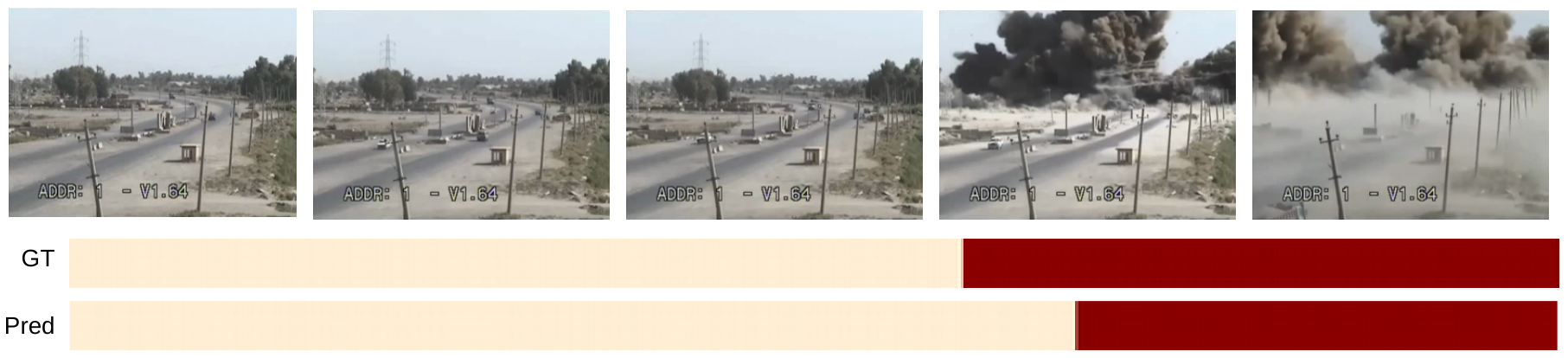} \\
\includegraphics[width=\linewidth]{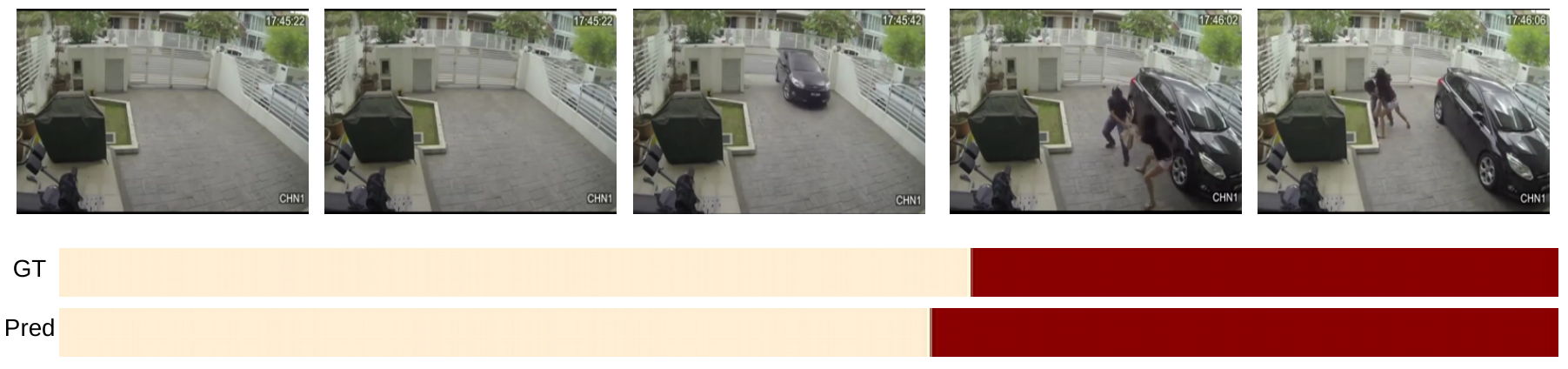} \\
\end{tabular}
\caption{Ground truth (GT) and prediction (pred) timelines (a) from Explosion category (b) from Robbery category. Red presents abnormal classes.}
\label{fig:timeline}
\end{center}
\end{figure}

In this paper, we propose ADNet, an anomaly detection network to localize anomaly events in temporal space of videos. ADNet uses temporal convolutions to localize abnormal cases in temporal space. While other temporal convolutional networks \cite{mstcn, tcn} require to get a whole video to produce a result, ADNet utilizes a sliding window approach and gets clips of a video in sequential order. Thus, varying sized videos can be handled effectively and loss of information due to fitting of the whole video to a fixed-size input can be prevented. Window based anomaly localization provides better performance when determining anomalies in videos and also enables real-time processing of surveillance videos and provides an indirect data augmentation mechanism to increase training data. After extracting features from each video clip by using a spatio-temporal deep network, ADNet takes features of all clips in a window and produce a separate anomaly score for each clip. We further define AD Loss function to increase anomaly detection accuracy. When ADNet is trained with the proposed AD Loss, performance of detecting abnormal segments increases.

We also propose to use segment based F1@k metric to measure effectiveness of an anomaly detection model instead of clip based AUC metric. AUC is not a good metric for this purpose since it does not take into account temporal order of clips. It penalizes minor shifts in temporal segments and can not effectively punish short false positive segments in temporal space. F1@k is calculated on IoU with k percentage of temporal segments obtained by thresholding anomaly scores of clips with 0.5 value. It is better in terms of measuring how correctly predicted temporal segment matches with ground truth temporal segment. 

As a last contribution we add two more anomaly classes to UCF Crime \cite{sultani2018real} dataset, which consists of 13 anomaly classes. Fig \ref{fig:timeline} shows ground truth and prediction timelines for two different anomaly classes. UCF Crime \cite{sultani2018real} data set provides temporal anomaly annotations in the test set, but only provides video level annotations in the training set. We provide temporal annotations for all anomaly classes of the data set. Finally, we thoroughly evaluate our model on UCF Crime dataset and extended dataset version, UCF Crime V2. ADNet produces promising abnormal segment detection score with \textbf{28.32 F1@10} while the baseline model's \cite{sultani2018real} score is \textbf{4.13 F1@10}. For the normal and abnormal segments, ADNet has 58.16 F1@10 score while baseline model's score is 45.20 F1@10.

\section{Related Works}
This work focuses on anomaly detection but it also utilizes action localization and recognition approaches. Thus, this section outlines the related works on anomaly detection, action recognition, and temporal action localization fields.

\textbf{Anomaly Detection} aims to detect events which are highly deviates from normal events. \citet{kratz2009anomaly} propose a statistical approach to  detect anomalies in extremely crowd scenes. \citet{mahadevan2010anomaly} propose a framework and a dataset to detect spatial and temporal anomalies in crowded scenes by using Gaussian Mixture Model on temporal space and discriminant saliency on spatial space. \citet{li2013anomaly} studies localization of anomaly on the same dataset. \citet{yuan2014online} propose structural context  descriptor (SCD) and 3D discrete cosine transform (DCT) multi-object tracker to detect and localize anomalies in crowded scenes. \citet{sultani2018real} propose model which is our baseline and UCF Crime dataset that has videos annotated with normal and abnormal labels. Baseline model has trained with multiple instance learning (MIL) and special loss on weakly supervised UCF Crime dataset to predict clip wise label. Videos are split to 32 segments. Each segment has extracted features, which are average of extracted clip features in the segment. Positive and negative bags consist of positive and negative segments, respectively. The model is trained with proposed loss. \citet{shah2018cadp} propose CADP dataset and a method for traffic anomaly detection. \cite{huang2020intelligent, yao2019unsupervised} studies anomaly detection of  vehicles from traffic videos or streams.

\textbf{Action Recognition} tries to classify actions on trimmed videos. \citet{c3d} proposes Convolutional 3D (C3D) network which has 3D convolutions to capture spatio-temporal features. \citet{i3d} propose Two-Stream Inflated 3D (I3D) network to learn better temporal features with different architecture by using 3D convolutions. I3D has two sub networks as \cite{simonyan2014two}, where they accept RGB and Optic Flow inputs. Classification probabilities are averaged before final prediction. ADNet uses only RGB branch of I3D to work more efficiently. Temporal Shift Module (TSM) \citet{tsm} propose to  predict action classes with pseudo-3D convolutions efficiently. Shifting some channels in one ore two direction of ResNet \cite{resnet}, Temporal Segment Networks (TSN) \cite{wang2016temporal}, provides temporal information to classify actions on Kinetics \cite{kinetics} and Something-Something\cite{goyal2017something} datasets. In order to decrease FLOPs, \cite{tran2018closer, qiu2017learning, sun2015human} use pseudo-3D convolution that is combination of 2D convolution in spatial space and 1D convolution in temporal space. \citet{tran2019video} propose Channel Separated Network (CSN), which decreases FLOPs by using 3D group convolutions instead of 3D convolutions.

\textbf{Temporal Action Localization} is that detecting and segmenting actions of untrimmed videos in temporal space. \citet{mstcn} proposes Temporal Convolutional Networks (TCN) by inspiration of a speech synthesis method, WaveNet \cite{wavenet}. 1D convolutions are applied on temporal space of extracted features of consecutive frames. Encoder-Decoder TCN (ED-TCN) achieves 68 F1@10 score, while Dilated TCN achieves 52.2 F1@10 score on 50 Salads \cite{50salads} dataset. ED-TCN and Dilated TCN accept features from VGG-style \cite{vgg} spatial networks. \citet{iqbal2019enhancing} proposes improving temporal action localization via transfer learning from I3D action recognition network \cite{i3d} with BoW-Network. BoW-Network creates visual words, which has 4000 components with k-means algorithm on Thumos \cite{idrees2017thumos} dataset. \citet{mstcn} proposes Multi-Stage TCN (MS-TCN) network and a loss function, which is combination of classification and smoothing loss.  MS-TCN achieves 76.3 F1@10 score on 50 Salads \cite{50salads} dataset.  \citet{ding2018weakly} propose Temporal Convolutional Feature Pyramid Network (TCFPN) and Iterative Soft Boundary Assignment (ISBA). ISBA generates temporal segments of given actions without temporal boundaries by using transcripts of video in Hollwood Extended \cite{bojanowski2014weakly} dataset. TCFPN is trained with weakly-supervised data as in \cite{shou2018autoloc, nguyen2018weakly}. \citet{gao2018ctap} uses sliding windows with different sizes to predict action scores at CTAP, collected windows are fed to temporal convolutional network.  \citet{buch2017sst} proposes SST that uses GRU instead of sliding window approch,  the GRU accepts C3D clip embeddings as input.

\section{Method}\label{section:method}

\begin{figure*}
\centering
\begin{subfigure}[b]{0.49\textwidth}
\centering
\includegraphics[width=0.8\linewidth]{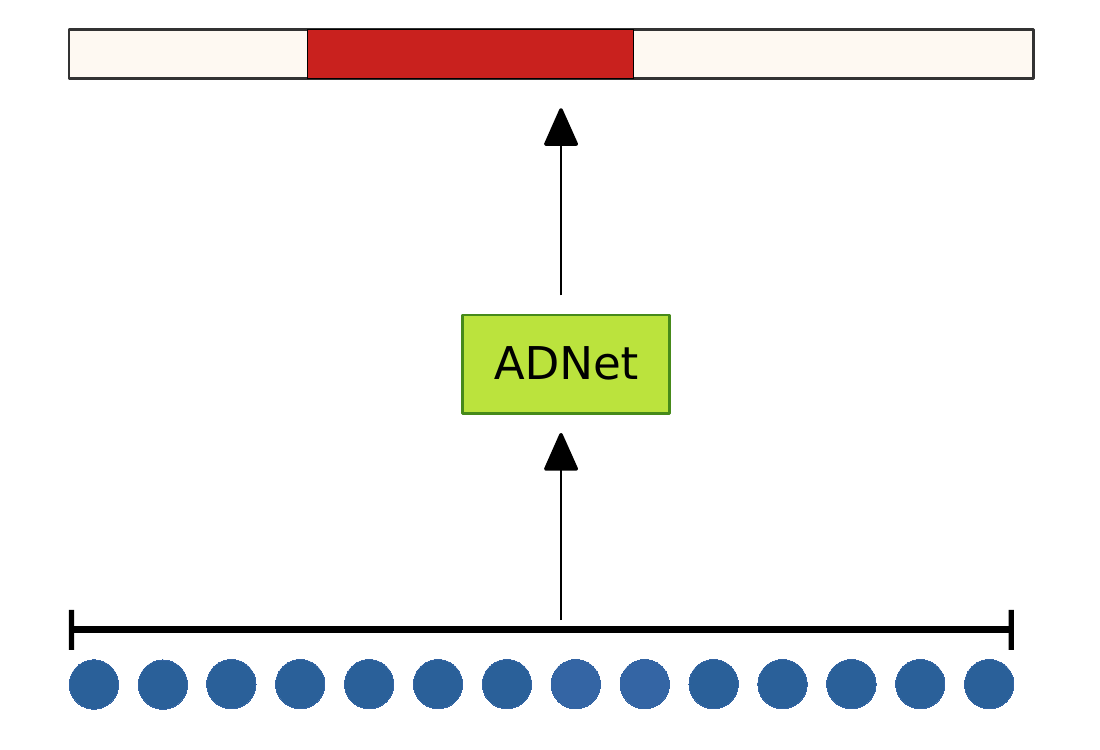}
  \caption{}
  \label{fig:temporal-window-a}
 \end{subfigure}
      \hfill
\begin{subfigure}[b]{0.49\textwidth}
\includegraphics[width=\linewidth]{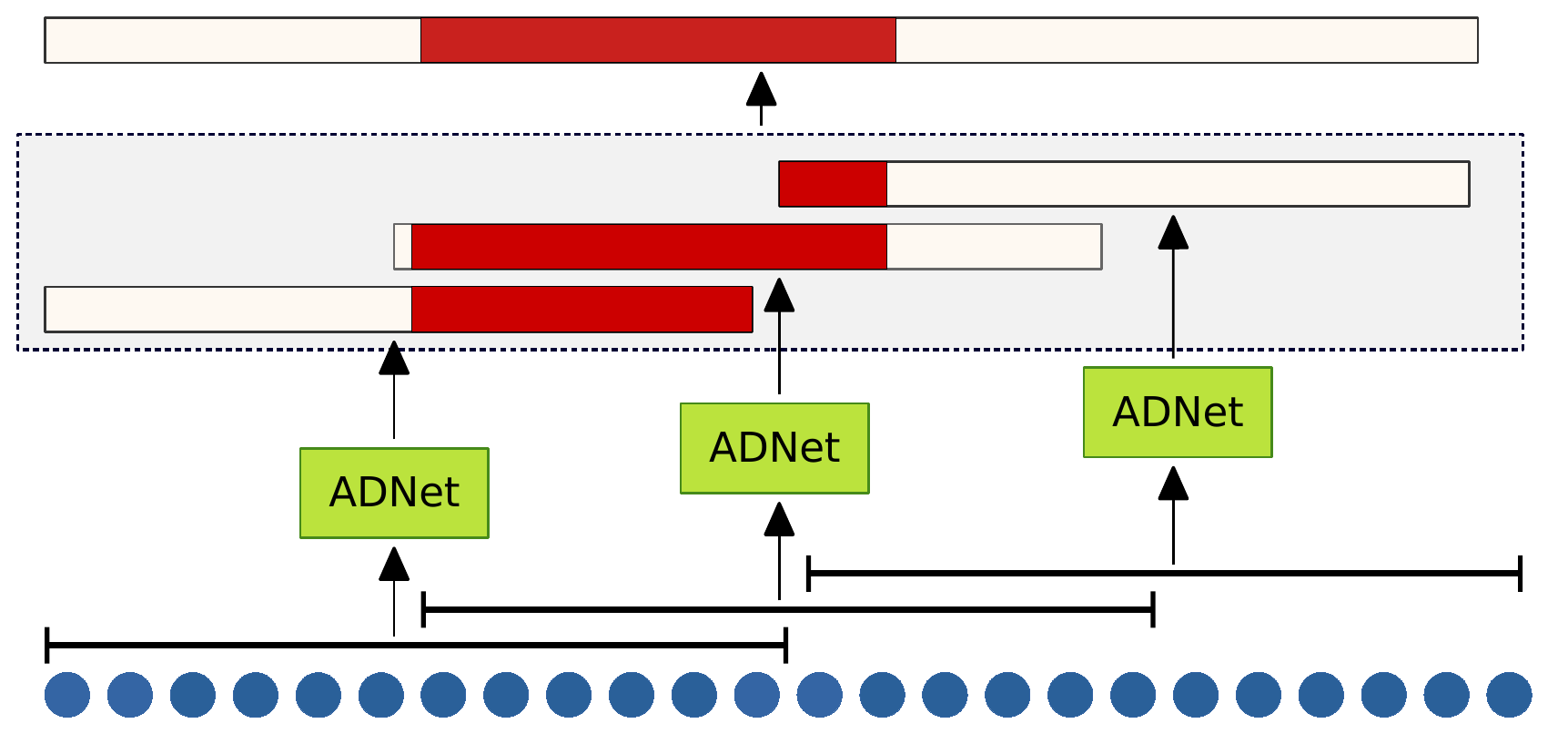}
  \caption{}
  \label{fig:temporal-window-b}
 \end{subfigure}
   \caption{Temporal convolutional networks \textit{(a)} accepts clip features $x_i$ which are presented with blue circles. Instead of convolving over clip features of a video as in \textit{(a)}, clip features are divided to equal windows to pass ADNet \textit{(b)}. Windows intersect with previous windows, output scores in overlapped parts are averaged to get final output scores. Abnormal results are presented with red colour.}
  \label{fig:temporal-window}

\end{figure*}

In this section we introduce Anomaly Detection Network (ADNet) for anomaly detection and segmentation in temporal space. Input to ADNet is a set of features  $x_{1:C}=(x_1, x_2, ..., x_C)$ extracted using a spatio-temporal CNN for each clip of given video, where $C$ is number of clips for a video. A clip consists of a number of  consecutive frames of video. ADNet outputs anomaly probabilities $y_{1:C}=(y_1, y_2, ... , y_C)$ for each clip, where  $y_t  \in  [0, 1]$. Target values are $a_{1:C}=(a_1, a_2, ... , a_C)$ where $a \in \{0, 1\}$. 0 and 1 indicates normal and abnormal classes. Labels are generated from output probabilities by applying  0.5 threshold value, which can be changed in inference time to adjust the model for different anomaly conditions. 

We describe ADNet in Section \ref{sec:adnet-method}. Then we describe Temporal Sliding Window in Section \ref{sec:window-method}. At the end of method section, we discuss the proposed loss in Section \ref{sec:loss-method}.

\subsection{Anomaly Detection Network} \label{sec:adnet-method}
To predict anomalies in a timeline by using temporal information, we adapt and improve MS-TCN \cite{mstcn} model which makes classification and temporal segmentation. We input all clip features of a video without splitting to small parts.

We define a sequence of stages, which has a series of blocks that consists of convolutional layers. Each stage outputs anomaly scores, $y_t$, for each clip features, $x_t$. Input of next stage is the output of previous stage, except the first stage. Input features are shrink with $D^H$ 1x1 convolutions on $D^0$ channels, where $D^0$ is the dimension of an input feature and $D^H$ is the number of channels of features outputted by hidden layers of stages. Let $V^{HxT}$ be output of each layer except the last layer in a stage. Dilated convolutions in layers are convolved on temporal dimension of $V^{HxT}$. Dilation rate of convolution is calculated with $2^l$ formula, where $l$ is order of layer in stage (i.e. $0, 1, 2, .., L$). Dilated convolutions increase receptive field of network with small kernel size. The activation function applied after convolutions is ReLU. Thus, 1x1 convolution follows ReLU. Residual connections between before and after convolutions in a block avoids gradient vanishing problem. Output of a stage is defined as follows:
\begin{equation}
    Y^s = Sigmoid(W*V + b)
\end{equation}
where $*$ is the convolution operation, $W$ is filter kernel and $b$ is bias value of kernel, $Y^s \in \mathbb{R}^{1xT}$ is the vector of anomaly scores outputted by $s_{th}$ stage.

We input the output previous stage, $Y^{s-1}$, to the next stage, $Y^s$. At the start of the stage, 1x1 convolution is applied to increase channel size from 1 to $D^H$.

\subsection{Temporal Sliding Window} \label{sec:window-method}

In the previous section, we input the whole clip features $x_{1,..,T}$ to the network. In order to provide generalization by data augmentation we split consecutive clip features to windows, where each window contains $W$ consecutive clip features, $(x_1, x_2, ..., x_W)$. If video has less clip features than window width, $T < W$, we pad window with empty clip features $x_0$, which filled with 0 values. Let $x=(x_1, x_2, x_0, ..., x_0)$  be an input and $m=(1, 1, 0, ... 0)$ be the mask for $x$, where $x \in \mathbb{R}^{D_0xW}$ and $m \in \mathbb{R}^{1xW}$. Information flow from padded empty clip features is blocked by masking outputs of convolutions as follows:
\begin{equation}
    V^{l}=(W_2 * ReLU(W_1 * V^{l-1} + b_1 ) + b_2) \cdot m
\end{equation}

where $V^l$ is the output of $l^{th}$ layer, $(W_1, W_2, b_1, b_2)$ are parameters of convolution, $\cdot$ is dot product.

This makes also online anomaly detection and segmentation possible, while they are impossible in the network accept whole clip features as input. Window stride size is set to half of the window size to obtain smoother results in inference, while increasing number of training clip windows. In other words, we augment data with both of splitting videos to windows and half stride windows during training. Start position of window $i$ is calculated with $w_i^{start} = W/2 * (i-1)$, end position formula is $w_i^{end} = W + W/2*(i-1)$. We use same window size for a network in both train and test stages. In the inference, we average anomaly scores of overlapping windows, as in Figure \ref{fig:temporal-window}.

Each layer in a stage except the last layer outputs $V^{TxD_H}$, where dimension of output is same for the convolutions. To match input dimension with output dimension, we pad input $P$, where $P=\floor*{K / 2}*2^L$, $P<W$, $K$ is kernel size. If padding  $P$ is equal to input size of window $W$, more than half of the inputs of convolution would be padding elements. Window width can not be more than the receptive field, where $ReceptiveField=2^{l+1}-1$. Therefore, to avoid information loss, we determine maximum number of layers for a window width $W$ as follows:

  \begin{equation}\label{formula:numlayers}
    L = \ceil*{log_2{\frac{W}{\floor*{K / 2}}}}
  \end{equation}

\subsection{Loss} \label{sec:loss-method}

\begin{figure}[h]
\centering
\includegraphics[width=0.7\linewidth]{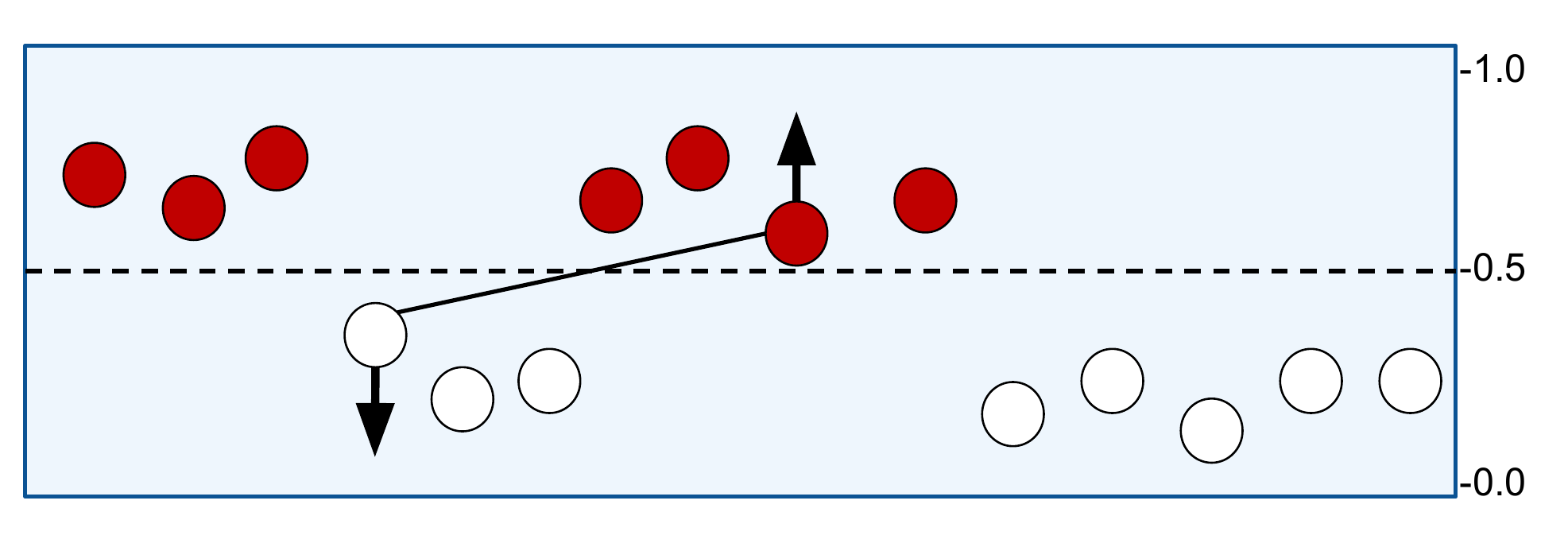}
  \caption{AD Loss increases distance between nearest opposite pair in manner of anomaly score for each clip. Red circles represent abnormal clips, white circles represent normal clips}
  \label{fig:loss}
\end{figure}

Intuition behind the loss is similar to VSE++ \cite{vse++}, increasing distance between hard pairs which are closest wrong embeddings in terms of cosine distance. We define hard pair as clips from opponent classes which have closest anomaly score outputs as in Figure \ref{fig:loss}. Hard pairs change from in each of step of training. Combination of $L_{MSE}$ mean squared error loss and $L_{AD}$ anomaly detection loss produces the final loss value as in Formula \ref{formula:loss}. Contribution of $L_{AD}$ is controlled with $0<\lambda<1$ parameter in Formula \ref{formula:loss}. $L_{AD}$ is calculated as in Formula \ref{formula:anomaly-loss}, where $y_{A}$ is score of abnormal input, $y_{HN}$ hard normal of the abnormal input, $y_{N}$ is score of normal input, and $y_{HA}$ hard abnormal of the normal input. $\alpha$ parameter controls ideal distance between hard pairs.
    \begin{equation}\label{formula:loss}
        L_s = L_{MSE} + \lambda * L_{AD}
    \end{equation}
    \begin{equation}\label{formula:anomaly-loss}
        L_{AD} = max( -(y_A - y_{HN} - \alpha) - (y_{HA} - y_N - \alpha), 0)
    \end{equation}
As we mentioned in Section \ref{sec:adnet-method}, each stage outputs anomaly probability for each input clip feature of given video. We calculate loss for each stage output as in Formula \ref{formula:stageloss}. The summation of losses are minimized in training.
    \begin{equation}\label{formula:stageloss}
        L = \sum_{s}{L_s}
    \end{equation}{}

\subsection {Implementation Details}
We extract clip features from I3D\cite{i3d} by applying average pooling to activations before classification layer. Video clips for I3D are generated with 16 frame temporal slide. We chose 16 temporal slide instead of 1 to decrease inference time. TSM\cite{tsm} is second feature extractor used in our study. we use Adam \cite{adam} optimizer with 5e-4 learning rate in all experiments . We trained and tested our models on PyTorch \cite{pytorch} framework. Kernel size and channel size of ADNet are 3 and 64 respectively in all settings. Labels of clips are determined by distribution of normal and abnormal classes.

\section{Experiments}
\subsection{Evaluation Metrics} 
\begin{figure}[H]
\centering
\includegraphics[width=0.95\linewidth]{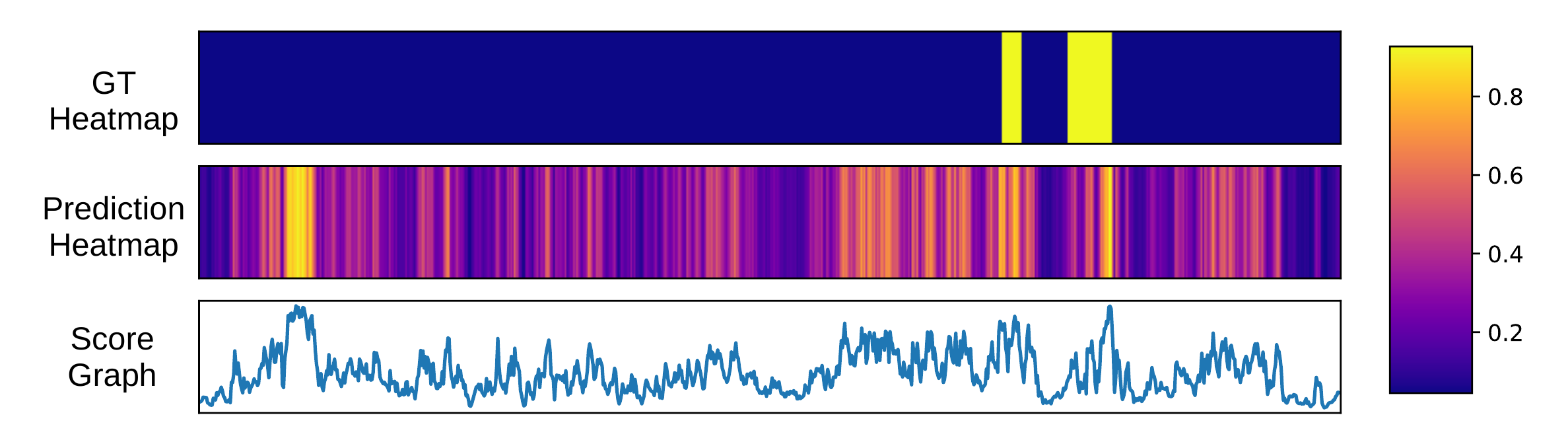}
  \caption{First two rows are heatmaps of ground truth, predicted anomaly scores for \textit{Assault 10} video from the test set. Last row is graph of anomaly scores which is between 0 and 1. AUC score is 74.83, F1@25 score is 24.99}
  \label{fig:auc-vs-f1}
\end{figure}

The baseline method \cite{sultani2018real} and most other studies use AUC metric to measure anomaly detection performance. Abnormal cases happen in a segment of timeline. AUC metric evaluates performance of each clip independently, in other words, avoiding the temporal orders of the clips. For this reason AUC cannot do the necessary punishment when short false positives arises. An appropriate metric should not penalize minor shifts between predicted segments and ground truth segments while penalizing over-segmentation errors. However, clip wise AUC metric penalizes minor shifts. Therefore, we approach to the problem as temporal action localization. We adapt F1@k metric proposed in \cite{tcn} for evaluating anomaly detection performance, which overcomes the weaknesses of AUC metric. 

Abnormal segments form small part of ground truth in a test sample. For this reason, either wrong predictions in abnormal segments or small false positives in normal segments do not sufficiently affect AUC score.  Figure \ref{fig:auc-vs-f1} shows timelines of ground truth and prediction segmentation for a test video. While AUC score of the test video is \textbf{74.83} , F1@25 score is \textbf{24.99} . This example shows the robustness of F1@k metric comparing to AUC metric for this problem. 
 
F1@k is calculated by k percentage intersection over union (IOU) between predicted temporal segments and ground truth temporal segments. However, \cite{tcn} does not include background segments to F1@k metric evaluation. In order to increase penalization for over-segmentation of abnormal segments, we do not consider segments with normal events as background segments, we include them to evaluation.
 
As mentioned before, we extract features for each clip from spatio-temporal networks. Let $C_i=(f_{n*i}, ..., f_{n*(i+1)}) $ be a clip where $f_j$ is $j_{th}$ frame of given video. Since accepted number of frames from feature extractors can be different, number of clips can be different for given video. To evaluate fairly, we produce frame level labels or scores from clip level results by copying clip scores $y_i$ to frame scores $\hat{y}_j$ of the clip, $\hat{y}_{n*i, ..., n*(i+1)} = y_i$. We make evaluation in frame-level instead of clip level.

\subsection{Dataset}

For the evaluation of the model, we use UCF Crime data set \cite{sultani2018real}, which consists of 13 anomaly classes. We have added two different anomaly classes to the data set, which are "molotov bomb" and "protest" classes. We also have added 33 videos to fighting class. In total, we have added  216 videos to the training set, 17 videos to the test set. Test set of UCF Crime data set has temporal annotations and classifications. However, training videos of UCF Crime data set are classified in video-level and temporal annotations are not provided for the training set. To train models with temporal information, we annotated anomalies of training videos in temporal domain. In order to annotate efficiently, annotators have used seconds as basis and assumed that the frames in a second all belong to the same class. Since we extend the dataset with new anomaly classes and temporal annotations for training videos, we name new version of the dataset as UCF Crime V2. 

Since baseline model has been trained without new anomaly classes, we do not evaluate the baseline on UCF Crime v2 but we evaluate on UCF Crime v1. We investigate effects of window size, number of layer, feature extractor and loss functions on UCF Crime v2 dataset. 

\subsection{Results}
In this section we present and discuss results of evaluation experiments.

\begin{table}
\parbox{.45\linewidth}{
\centering
         \begin{tabular}{ c | c  c  c } 
        \Xhline{1.3pt}
          \multirow{2}{*}{\parbox{1.5cm}{Methods}}
         & \multicolumn{3}{|c}{ \thead{UCF Crime V2}} \\
         \cline{2-4}
          & \thead{F1@10} & \thead{F1@25} & \thead{F1@50}  \\ [0.5ex] 
        \Xhline{1.0pt}
         \thead{ADNet W64-S5-L8} & 53.03 & 47.03 & 33.95 \\
         \hline
         \thead{ADNet W64-S5-L7} & 53.82 & 48.24 & 34.02 \\
         \hline
         \thead{ADNet W64-S5-L6} & \textbf{58.73} & \textbf{52.38} & \textbf{40.98}  \\
         \hline
         \thead{ADNet W64-S5-L5} & 48.12 & 39.95 & 29.10 \\
         \hline
         \thead{ADNet W64-S5-L4} & 51.35 & 42.19 & 28.17 \\
        \Xhline{1.3pt}
        \end{tabular}
        \caption{Comparison of number of layers. I3D is feature extractor in the experiments.}
    \label{table:layer-results}
}
\hfill
\parbox{.47\linewidth}{
         \begin{tabular}{  c | c  c  c   } 
         \Xhline{1.3pt}
          \multirow{2}{*}{\parbox{1.5cm}{Methods}} 
         & \multicolumn{3}{|c}{ \thead{UCF Crime V2}} \\
         \cline{2-4}
          &\thead{F1@10} & \thead{F1@25} & \thead{F1@50}  \\ [0.5ex] 
         \Xhline{1.0pt}
         \thead{ADNet W32-S3-L5 } & 50.78 & 43.22 & 32.05 \\
         \hline
         \thead{ADNet W64-S5-L6 } &  \textbf{58.73} & \textbf{52.38} & \textbf{40.98} \\
         \hline
         \thead{ADNet W128-S8-L7 } & 55.61 & 50.04 & 36.60 \\
        \hline
        \thead{ADNet w/o Window } & 51.62 & 43.86 & 32.30 \\
        \Xhline{1.3pt}
        \end{tabular}
        \caption{Comparison of different window sizes (W: window, S: number of stages, L: number of layers in a stage). I3D is feature extractor in the experiments. }
    \label{table:window-results}
}
\end{table}

\textbf{Effect of Temporal Sliding Window.} We start evaluation by showing the effect of temporal sliding window method. We compare different window widths and without window of ADNet in Table \ref{table:window-results}. Experiments in the table have been made on UCF Crime V2 dataset. Number of layers in this experiment is set according to window width and Formula \ref{formula:numlayers}. Window size and number of stages might have different values, which are experimented in our ablation study. Table \ref{table:window-results} presents three window sizes: 32, 64 and 128 clips in a window. Number of stages for each window width are selected by taking into account the best result of that window. Generally, number of stages increases in parallel to window width. As we mentioned before, temporal sliding window augments data during training. Decreasing window size provides better augmentation, but temporal information is collapsed. There is a trade-off between data augmentation and temporal knowledge. Table \ref{table:window-results} shows that the best result is achieved with 64 temporal window width. All of the temporal sliding window results are better than straight forward ADNet (w/o Window) in terms of F1@k scores. This means that temporal sliding window improves performance of ADNet in UCF Crime v2 dataset. 

\textbf{Effect of Number of Layers.} We have compared window widths where number of layers is calculated with Formula \ref{formula:numlayers}. In this part we discuss effect of number of layers to performance. Table \ref{table:layer-results} shows results of experiments with fixed window width and number of stages. ADNet with more layers achieves better results. However, a layer number more than the maximum layer number calculated by Formula \ref{formula:numlayers} cause information loss as mentioned in section \ref{section:method}.

\begin{table}
    \begin{center}
         \begin{tabular}{ c | c  c || c  c } 
        \Xhline{1.3pt}
          \multirow{3}{*}{\parbox{1.5cm}{Methods}} & 
         \multicolumn{4}{c}{ \thead{UCF Crime V2}} \\
         
         \cline{2-5} &\multicolumn{2}{ c ||}{\thead{Abnormal Segments}} &\multicolumn{2}{c  }{\thead{Normal Segments}} \\
         
         \cline{2-5} & \thead{F1@10} & \thead{F1@25} & \thead{F1@10} & \thead{F1@25} \\ [0.5ex] 
        \Xhline{1.0pt}
         \thead{ADNet (MSE)} & 29.00 & 19.33 & \textbf{71.23} & \textbf{66.44}  \\
        \hline
         \thead{ADNet (MSE+AD)} & \textbf{32.16} & \textbf{20.70} & 56.34 & 50.54 \\
        \Xhline{1.3pt}
        \end{tabular}
        \caption{Comparison of loss functions. I3D is feature extractor in the experiments.}
    \label{table:mse-ad-loss}
    \end{center}
\end{table}

\textbf{Loss function.} We propose the AD loss function, which tries to maximize distance between hard pairs from opponent classes. Table \ref{table:mse-ad-loss} shows results for MSE loss and MSE and anomaly detection (AD) loss. In these experiments, window width is 64, number of stages is 5, and number of layers is 6. The parameters are selected based on the previous experiments. $\lambda$ parameter in Formula \ref{formula:anomaly-loss} controls contribution of AD loss to total loss, $\lambda$ is set to 0.5 in these experiment. The results show that MSE+AD loss is more successful than MSE loss at abnormal segments, but MSE+AD loss is worse than MSE loss at normal segments. For this reason we use MSE loss only in other experiments.

\begin{table}
    \begin{center}
         \begin{tabular}{ c |  c  c   || c  c } 
        \Xhline{1.3pt}
          \multirow{3}{*}{\parbox{1.5cm}{Methods}} & 
         \multicolumn{4}{c}{ \thead{UCF Crime V2}} \\
         
         \cline{2-5} &\multicolumn{2}{ c ||}{\thead{Abnormal Segments}} &\multicolumn{2}{c  }{\thead{Normal Segments}} \\
         
         \cline{2-5}
          & \thead{F1@10} & \thead{F1@25} & \thead{F1@10} &  \thead{F1@25} \\ [0.5ex] 
        \Xhline{1.0pt}
         \thead{ADNet [I3D]} & 29.0 & 19.33 & \textbf{71.23} & \textbf{66.44} \\
        \hline
         \thead{ADNet [TSM]} &\textbf{ 33.50 }& \textbf{22.78} & 61.71 & 57.52 \\
        \Xhline{1.3pt}
        \end{tabular}
        \caption{Comparison of different spatio-temporal feature extractors}
    \label{table:i3d-tsm}
    \end{center}
\end{table}

\textbf{Feature Extractor.} We have experimented on two different spatio-temporal feature extractors. In the previous experiments, we used I3D \cite{i3d} as the feature extractor. Temporal Shift Module (TSM) \cite{tsm} is more a efficient action recognition network than I3D network while performance on Kinetics \cite{kinetics} dataset is similar. To extract features from video clips,  we have used TSM as an alternative to I3D. Table \ref{table:i3d-tsm} shows the results for I3D and TSM networks, where parameters of ADNet are as follows, window width is 64, number of stages is 5 and number of layers is 6. Although I3D and TSM get different sized input clips and produce outputs in different formats, this experiment shows us that ADNet can utilize different feature extractor networks. As a result of this experiment, we observed that I3D features is more useful in ADNet network for segmenting normal events while TSM features is better for segmenting abnormal events.

\begin{table*}
    \begin{center}
         \begin{tabular}{ c | c  c  c ||  c  c  c || c c c } 
        \Xhline{1.3pt}
          \multirow{3}{*}{\parbox{1.5cm}{Methods}} & 
          \multicolumn{9}{c}{ \thead{UCF Crime v1}} \\
         \cline{2-10} & \multicolumn{3}{c||}{ \thead{Abnormal Segments}} 
         & \multicolumn{3}{ c ||}{ \thead{Normal Segments}} & \multicolumn{3}{ c }{ \thead{All Segments}} \\
         \cline{2-10}
          & \thead{F1@10} & \thead{F1@25} & \thead{F1@50} & \thead{F1@10} & \thead{F1@25} & \thead{F1@50} & \thead{F1@10} & \thead{F1@25} & \thead{F1@50}   \\ [0.5ex] 
        \Xhline{1.0pt}
         \thead{Baseline \\ Network [C3D]} & 4.13 & 1.65 & 0 & 63.27 & 56.36 & 46.54 & 45.20 & 39.64 & 32.32 \\ 
         \hline
         \thead{MLP [C3D]} & 7.34 & 1.86 & 0.65 & 65.96 & 63.15 & 54.75 & 49.40 & 44.83 & 38.15 \\
         \hline
         \thead{ED-TCN [I3D]} & 21.18 & 12.63 & 4.88 & 61.60 & 53.59 & 36.71 & 47.81 & 39.61 & 25.85  \\
         \hline
         \thead{ADNet [I3D] (Ours)} & \textbf{28.32} & \textbf{18.71} & \textbf{9.44} & \textbf{71.23} & \textbf{66.44} & \textbf{55.48}  & \textbf{58.16} & \textbf{51.85} & \textbf{41.29 } \\
        \Xhline{1.3pt}
        \end{tabular}
        \caption{Performance comparison of state-of-the art methods}
    \label{table:results}
    \end{center}
\end{table*}

\textbf{Comparison with State-of-the-Arts.} Temporal annotations of training set of UCF Crime data set has not been available until our study. Baseline model \cite{sultani2018real} has not been trained on temporarily annotated train set. In order to compare ADNet with models trained on temporarily annotated training set, we have trained two models which accept features extracted from I3D network. First model is Multi Layer Perceptron (MLP) with 3 layers as in baseline network \cite{sultani2018real} which generates predictions clip-wise. Second model is Encoder Decoder Temporal Convolutional Network (ED-TCN) proposed in \cite{tcn} with temporal sliding window method as in Section \ref{sec:window-method}. According to the results presented in Table \ref{table:results}, our proposed model achieves better scores than baseline network and other models in all categories on UCF Crime v1 test set. While baseline model achieves \textbf{4.13 F1@10} score at abnormal segments, our model achieves \textbf{28.32 F1@10} score. Window width, number of stages and number of layers of ADNet in Table \ref{table:results} are 64, 5, and 6, respectively. Ground truths and predictions in timeline of two test videos from Explosion and Robbery categories are in Figure \ref{fig:timeline}.

\section{Conclusion}
We proposed a temporal anomaly detection network, which enables to localize anomalies in videos with temporal convolutions. In our knowledge, this is the first approach to formulate anomaly detection problem in a similar way to action localization problem. We also introduced AD loss function, which enabled to have better detection performance in abnormal classes. We evaluated and discussed the effects of model parameters, which are window width, number of layers, number of stages, feature extractor, and loss functions. We also extended the UCF Crime anomaly dataset with two additional anomaly classes and temporal annotations of training videos. Extensive evaluations of the model shows that the model has promising results on real world anomaly videos. Window based operation of the model allows processing of online video streams. We also investigated the evaluation metrics in terms of measuring anomaly detection performance. Since F1@k does not penalize minor shifts and does punish short false positive temporal segment predictions, we concluded that F1@k metric is better than AUC metric for measuring anomaly detection performance. In the future works, scene context information and relations between objects can be utilized to improve anomaly detection performance.

\subsection{Acknowledgements}
This work was supported in part by the Scientific and Technological Research Council of Turkey (TUBITAK) under Grant No. 114G028 and 1198E098.  







%
%
%
%
\bibliographystyle{plainnat}
\bibliography{samplepaper}

\begin{thebibliography}{36}
\providecommand{\natexlab}[1]{#1}
\providecommand{\url}[1]{\texttt{#1}}
\expandafter\ifx\csname urlstyle\endcsname\relax
  \providecommand{\doi}[1]{doi: #1}\else
  \providecommand{\doi}{doi: \begingroup \urlstyle{rm}\Url}\fi

\bibitem[Bojanowski et~al.(2014)Bojanowski, Lajugie, Bach, Laptev, Ponce,
  Schmid, and Sivic]{bojanowski2014weakly}
Piotr Bojanowski, R{\'e}mi Lajugie, Francis Bach, Ivan Laptev, Jean Ponce,
  Cordelia Schmid, and Josef Sivic.
\newblock Weakly supervised action labeling in videos under ordering
  constraints.
\newblock In \emph{European Conference on Computer Vision}, pages 628--643.
  Springer, 2014.

\bibitem[Buch et~al.(2017)Buch, Escorcia, Shen, Ghanem, and
  Carlos~Niebles]{buch2017sst}
Shyamal Buch, Victor Escorcia, Chuanqi Shen, Bernard Ghanem, and Juan
  Carlos~Niebles.
\newblock Sst: Single-stream temporal action proposals.
\newblock In \emph{Proceedings of the IEEE conference on Computer Vision and
  Pattern Recognition}, pages 2911--2920, 2017.

\bibitem[Carreira and Zisserman(2017)]{i3d}
Joao Carreira and Andrew Zisserman.
\newblock Quo vadis, action recognition? a new model and the kinetics dataset.
\newblock In \emph{proceedings of the IEEE Conference on Computer Vision and
  Pattern Recognition}, pages 6299--6308, 2017.

\bibitem[Ding and Xu(2018)]{ding2018weakly}
Li~Ding and Chenliang Xu.
\newblock Weakly-supervised action segmentation with iterative soft boundary
  assignment.
\newblock In \emph{Proceedings of the IEEE Conference on Computer Vision and
  Pattern Recognition}, pages 6508--6516, 2018.

\bibitem[Faghri et~al.(2017)Faghri, Fleet, Kiros, and Fidler]{vse++}
Fartash Faghri, David~J Fleet, Jamie~Ryan Kiros, and Sanja Fidler.
\newblock Vse++: Improving visual-semantic embeddings with hard negatives.
\newblock \emph{arXiv preprint arXiv:1707.05612}, 2017.

\bibitem[Farha and Gall(2019)]{mstcn}
Yazan~Abu Farha and Jurgen Gall.
\newblock Ms-tcn: Multi-stage temporal convolutional network for action
  segmentation.
\newblock In \emph{Proceedings of the IEEE Conference on Computer Vision and
  Pattern Recognition}, pages 3575--3584, 2019.

\bibitem[Gao et~al.(2018)Gao, Chen, and Nevatia]{gao2018ctap}
Jiyang Gao, Kan Chen, and Ram Nevatia.
\newblock Ctap: Complementary temporal action proposal generation.
\newblock In \emph{Proceedings of the European conference on computer vision
  (ECCV)}, pages 68--83, 2018.

\bibitem[Goyal et~al.(2017)Goyal, Kahou, Michalski, Materzynska, Westphal, Kim,
  Haenel, Fruend, Yianilos, Mueller-Freitag, et~al.]{goyal2017something}
Raghav Goyal, Samira~Ebrahimi Kahou, Vincent Michalski, Joanna Materzynska,
  Susanne Westphal, Heuna Kim, Valentin Haenel, Ingo Fruend, Peter Yianilos,
  Moritz Mueller-Freitag, et~al.
\newblock The" something something" video database for learning and evaluating
  visual common sense.
\newblock In \emph{ICCV}, volume~1, page~5, 2017.

\bibitem[He et~al.(2016)He, Zhang, Ren, and Sun]{resnet}
Kaiming He, Xiangyu Zhang, Shaoqing Ren, and Jian Sun.
\newblock Deep residual learning for image recognition.
\newblock In \emph{Proceedings of the IEEE conference on computer vision and
  pattern recognition}, pages 770--778, 2016.

\bibitem[Huang et~al.(2020)Huang, He, Rangarajan, and
  Ranka]{huang2020intelligent}
Xiaohui Huang, Pan He, Anand Rangarajan, and Sanjay Ranka.
\newblock Intelligent intersection: Two-stream convolutional networks for
  real-time near-accident detection in traffic video.
\newblock \emph{ACM Transactions on Spatial Algorithms and Systems (TSAS)},
  6\penalty0 (2):\penalty0 1--28, 2020.

\bibitem[Idrees et~al.(2017)Idrees, Zamir, Jiang, Gorban, Laptev, Sukthankar,
  and Shah]{idrees2017thumos}
Haroon Idrees, Amir~R Zamir, Yu-Gang Jiang, Alex Gorban, Ivan Laptev, Rahul
  Sukthankar, and Mubarak Shah.
\newblock The thumos challenge on action recognition for videos “in the
  wild”.
\newblock \emph{Computer Vision and Image Understanding}, 155:\penalty0 1--23,
  2017.

\bibitem[Iqbal et~al.(2019)Iqbal, Richard, and Gall]{iqbal2019enhancing}
Ahsan Iqbal, Alexander Richard, and Juergen Gall.
\newblock Enhancing temporal action localization with transfer learning from
  action recognition.
\newblock In \emph{Proceedings of the IEEE International Conference on Computer
  Vision Workshops}, pages 0--0, 2019.

\bibitem[Kay et~al.(2017)Kay, Carreira, Simonyan, Zhang, Hillier,
  Vijayanarasimhan, Viola, Green, Back, Natsev, et~al.]{kinetics}
Will Kay, Joao Carreira, Karen Simonyan, Brian Zhang, Chloe Hillier, Sudheendra
  Vijayanarasimhan, Fabio Viola, Tim Green, Trevor Back, Paul Natsev, et~al.
\newblock The kinetics human action video dataset.
\newblock \emph{arXiv preprint arXiv:1705.06950}, 2017.

\bibitem[Kingma and Ba(2014)]{adam}
Diederik~P Kingma and Jimmy Ba.
\newblock Adam: A method for stochastic optimization.
\newblock \emph{arXiv preprint arXiv:1412.6980}, 2014.

\bibitem[Kratz and Nishino(2009)]{kratz2009anomaly}
Louis Kratz and Ko~Nishino.
\newblock Anomaly detection in extremely crowded scenes using spatio-temporal
  motion pattern models.
\newblock In \emph{2009 IEEE Conference on Computer Vision and Pattern
  Recognition}, pages 1446--1453. IEEE, 2009.

\bibitem[Lea et~al.(2017)Lea, Flynn, Vidal, Reiter, and Hager]{tcn}
Colin Lea, Michael~D Flynn, Rene Vidal, Austin Reiter, and Gregory~D Hager.
\newblock Temporal convolutional networks for action segmentation and
  detection.
\newblock In \emph{proceedings of the IEEE Conference on Computer Vision and
  Pattern Recognition}, pages 156--165, 2017.

\bibitem[Li et~al.(2013)Li, Mahadevan, and Vasconcelos]{li2013anomaly}
Weixin Li, Vijay Mahadevan, and Nuno Vasconcelos.
\newblock Anomaly detection and localization in crowded scenes.
\newblock \emph{IEEE transactions on pattern analysis and machine
  intelligence}, 36\penalty0 (1):\penalty0 18--32, 2013.

\bibitem[Lin et~al.(2019)Lin, Gan, and Han]{tsm}
Ji~Lin, Chuang Gan, and Song Han.
\newblock Tsm: Temporal shift module for efficient video understanding.
\newblock In \emph{Proceedings of the IEEE International Conference on Computer
  Vision}, pages 7083--7093, 2019.

\bibitem[Mahadevan et~al.(2010)Mahadevan, Li, Bhalodia, and
  Vasconcelos]{mahadevan2010anomaly}
Vijay Mahadevan, Weixin Li, Viral Bhalodia, and Nuno Vasconcelos.
\newblock Anomaly detection in crowded scenes.
\newblock In \emph{2010 IEEE Computer Society Conference on Computer Vision and
  Pattern Recognition}, pages 1975--1981. IEEE, 2010.

\bibitem[Nguyen et~al.(2018)Nguyen, Liu, Prasad, and Han]{nguyen2018weakly}
Phuc Nguyen, Ting Liu, Gautam Prasad, and Bohyung Han.
\newblock Weakly supervised action localization by sparse temporal pooling
  network.
\newblock In \emph{Proceedings of the IEEE Conference on Computer Vision and
  Pattern Recognition}, pages 6752--6761, 2018.

\bibitem[Oord et~al.(2016)Oord, Dieleman, Zen, Simonyan, Vinyals, Graves,
  Kalchbrenner, Senior, and Kavukcuoglu]{wavenet}
Aaron van~den Oord, Sander Dieleman, Heiga Zen, Karen Simonyan, Oriol Vinyals,
  Alex Graves, Nal Kalchbrenner, Andrew Senior, and Koray Kavukcuoglu.
\newblock Wavenet: A generative model for raw audio.
\newblock \emph{arXiv preprint arXiv:1609.03499}, 2016.

\bibitem[Paszke et~al.(2019)Paszke, Gross, Massa, Lerer, Bradbury, Chanan,
  Killeen, Lin, Gimelshein, Antiga, et~al.]{pytorch}
Adam Paszke, Sam Gross, Francisco Massa, Adam Lerer, James Bradbury, Gregory
  Chanan, Trevor Killeen, Zeming Lin, Natalia Gimelshein, Luca Antiga, et~al.
\newblock Pytorch: An imperative style, high-performance deep learning library.
\newblock In \emph{Advances in Neural Information Processing Systems}, pages
  8024--8035, 2019.

\bibitem[Qiu et~al.(2017)Qiu, Yao, and Mei]{qiu2017learning}
Zhaofan Qiu, Ting Yao, and Tao Mei.
\newblock Learning spatio-temporal representation with pseudo-3d residual
  networks.
\newblock In \emph{proceedings of the IEEE International Conference on Computer
  Vision}, pages 5533--5541, 2017.

\bibitem[Shah et~al.(2018)Shah, Lamare, Nguyen-Anh, and
  Hauptmann]{shah2018cadp}
Ankit~Parag Shah, Jean-Bapstite Lamare, Tuan Nguyen-Anh, and Alexander
  Hauptmann.
\newblock Cadp: A novel dataset for cctv traffic camera based accident
  analysis.
\newblock In \emph{2018 15th IEEE International Conference on Advanced Video
  and Signal Based Surveillance (AVSS)}, pages 1--9. IEEE, 2018.

\bibitem[Shou et~al.(2018)Shou, Gao, Zhang, Miyazawa, and
  Chang]{shou2018autoloc}
Zheng Shou, Hang Gao, Lei Zhang, Kazuyuki Miyazawa, and Shih-Fu Chang.
\newblock Autoloc: Weakly-supervised temporal action localization in untrimmed
  videos.
\newblock In \emph{Proceedings of the European Conference on Computer Vision
  (ECCV)}, pages 154--171, 2018.

\bibitem[Simonyan and Zisserman(2014{\natexlab{a}})]{simonyan2014two}
Karen Simonyan and Andrew Zisserman.
\newblock Two-stream convolutional networks for action recognition in videos.
\newblock In \emph{Advances in neural information processing systems}, pages
  568--576, 2014{\natexlab{a}}.

\bibitem[Simonyan and Zisserman(2014{\natexlab{b}})]{vgg}
Karen Simonyan and Andrew Zisserman.
\newblock Very deep convolutional networks for large-scale image recognition.
\newblock \emph{arXiv preprint arXiv:1409.1556}, 2014{\natexlab{b}}.

\bibitem[Stein and McKenna(2013)]{50salads}
Sebastian Stein and Stephen~J McKenna.
\newblock Combining embedded accelerometers with computer vision for
  recognizing food preparation activities.
\newblock In \emph{Proceedings of the 2013 ACM international joint conference
  on Pervasive and ubiquitous computing}, pages 729--738, 2013.

\bibitem[Sultani et~al.(2018)Sultani, Chen, and Shah]{sultani2018real}
Waqas Sultani, Chen Chen, and Mubarak Shah.
\newblock Real-world anomaly detection in surveillance videos.
\newblock In \emph{Proceedings of the IEEE Conference on Computer Vision and
  Pattern Recognition}, pages 6479--6488, 2018.

\bibitem[Sun et~al.(2015)Sun, Jia, Yeung, and Shi]{sun2015human}
Lin Sun, Kui Jia, Dit-Yan Yeung, and Bertram~E Shi.
\newblock Human action recognition using factorized spatio-temporal
  convolutional networks.
\newblock In \emph{Proceedings of the IEEE international conference on computer
  vision}, pages 4597--4605, 2015.

\bibitem[Tran et~al.(2015)Tran, Bourdev, Fergus, Torresani, and Paluri]{c3d}
Du~Tran, Lubomir Bourdev, Rob Fergus, Lorenzo Torresani, and Manohar Paluri.
\newblock Learning spatiotemporal features with 3d convolutional networks.
\newblock In \emph{Proceedings of the IEEE international conference on computer
  vision}, pages 4489--4497, 2015.

\bibitem[Tran et~al.(2018)Tran, Wang, Torresani, Ray, LeCun, and
  Paluri]{tran2018closer}
Du~Tran, Heng Wang, Lorenzo Torresani, Jamie Ray, Yann LeCun, and Manohar
  Paluri.
\newblock A closer look at spatiotemporal convolutions for action recognition.
\newblock In \emph{Proceedings of the IEEE conference on Computer Vision and
  Pattern Recognition}, pages 6450--6459, 2018.

\bibitem[Tran et~al.(2019)Tran, Wang, Torresani, and Feiszli]{tran2019video}
Du~Tran, Heng Wang, Lorenzo Torresani, and Matt Feiszli.
\newblock Video classification with channel-separated convolutional networks.
\newblock In \emph{Proceedings of the IEEE International Conference on Computer
  Vision}, pages 5552--5561, 2019.

\bibitem[Wang et~al.(2016)Wang, Xiong, Wang, Qiao, Lin, Tang, and
  Van~Gool]{wang2016temporal}
Limin Wang, Yuanjun Xiong, Zhe Wang, Yu~Qiao, Dahua Lin, Xiaoou Tang, and Luc
  Van~Gool.
\newblock Temporal segment networks: Towards good practices for deep action
  recognition.
\newblock In \emph{European conference on computer vision}, pages 20--36.
  Springer, 2016.

\bibitem[Yao et~al.(2019)Yao, Xu, Wang, Crandall, and
  Atkins]{yao2019unsupervised}
Yu~Yao, Mingze Xu, Yuchen Wang, David~J Crandall, and Ella~M Atkins.
\newblock Unsupervised traffic accident detection in first-person videos.
\newblock \emph{arXiv preprint arXiv:1903.00618}, 2019.

\bibitem[Yuan et~al.(2014)Yuan, Fang, and Wang]{yuan2014online}
Yuan Yuan, Jianwu Fang, and Qi~Wang.
\newblock Online anomaly detection in crowd scenes via structure analysis.
\newblock \emph{IEEE transactions on cybernetics}, 45\penalty0 (3):\penalty0
  548--561, 2014.

\end{thebibliography}

\end{document}